# A Hybrid Science-Guided Machine Learning Approach for Modeling Chemical Processes: A Review

Niket Sharma and Y. A. Liu*, AspenTech Center of Excellence in Process System Engineering, Department of Chemical Engineering, Virginia Polytechnic Institute and State University, Blacksburg, Virginia 24061.    *Corresponding author: design@vt.edu; (540) 231-7800

**Abstract:**

*This study presents a broad perspective of hybrid process modeling combining the scientific knowledge and data analytics in bioprocessing and chemical engineering with a science-guided machine learning (SGML) approach. We divide the approach into two major categories: ML complements science, and science complements ML. We review the literature relating to the hybrid SGML approach, and propose a systematic classification of hybrid SGML models. For applying ML to improve science-based models, we present expositions of direct serial and parallel hybrid modeling and their combinations, inverse modeling, reduced-order modeling, quantifying uncertainty in the process and even discovering governing equations of the process model. For applying scientific principles to improve ML models, we discuss the science-guided design, learning and refinement. For each sub-category, we identify its requirements, strengths and limitations, together with their published and potential applications. We also present several examples to illustrate different hybrid SGML methodologies for modeling chemical processes.*

## 1 | INTRODUCTION

Modeling of many physiochemical systems requires detailed scientific knowledge of the system which is not always feasible for complex processes. We make some assumptions when modeling the system with first principles that ultimately leads to some knowledge gaps in describing the original system. Even for the systems where the scientific knowledge is sufficient to model the system and there is limited data to estimate the multiple parameters of a first-principles model. We often apply data-based models to study the systems where scientific data are available since they are more accurate in prediction. However, data-based/machine learning models are *black-box models* which can over-fit the data and also produce scientifically inconsistent results. For better accuracy, ML models also require more data which is not always feasible for many problems. Therefore, it is important to integrate science-based knowledge and data-based knowledge for an accurate and scientifically consistent prediction, which we will refer to as ***hybrid science- guided machine learning (SGML) approach.***

The most popular hybrid SGML approach that is being practiced in different fields of science is to combine a data-based ML model with a science-based first-principles model. However, there are more ways to combine scientific knowledge and data-based knowledge. In this work, we focus on both aspects of science complementing ML, and ML complementing science.

In our development of the hybrid SGML approach, we have benefited from two latest references. In their 2017 article, Karpatne et. al.[1] suggest the theory-guided data science as a new paradigm for scientific discovery from data. They classify the theory-guided data science methods into different categories, such as theory-guided design of models, initialization, theory-guided refinement of data science outputs, hybrid models of theory of data science, and augmenting theory-based models using data science. In their 2020 article, Willard et. al.[2] classify the integration of physics-based modeling with ML methodology according to the modeling objectives. The latter include, for example, improving the predictions beyond physical models, downscaling the complexity of physics-based models, generating data, quantifying uncertainty, and discovering governing equations of the data-based model.

The objective of this paper is to present a review and exposition of scientific and engineering literature relating to the hybrid SGML approach, and propose a systematic classification of hybrid SGML models focusing on both science complementing ML models, and ML complementing science-based models. Section 2 gives a review of the broad applications of hybrid SGML approach in bioprocessing and chemical engineering. As the number of reported methodologies and applications continues to rise significantly, it is hard for a person unfamiliar with the subject to identify the appropriate approach for a specific application. This leads to our key focus in Sections 3 to 5, beginning with a systematic classification and exposition of hybrid SGML methodologies in Section 3. Section 4 explains different categories of applying ML to complement science-based models, discuss their requirements, strengths and limitations, suggest potential areas of applications, and present illustrative examples from chemical manufacturing. Section 5 focuses on different categories of applying scientific principles to complement ML models, together with their requirements, strengths and limitations, as well as their potential applications and illustrative examples. Section describes the challenges and opportunities for hybrid SGML approach for modeling chemical processes. Section 7 summarizes our conclusions.

This work differentiates itself from several recent reviews of hybrid modeling in bioprocessing and chemical engineering through the following contributions: (1) presentation of a broader hybrid SGML methodology of integrating science-guided and data-based models, and not just the direct combinations of first-principles and ML models; (2) classification of the hybrid model applications according to their methodology and objectives, instead of their areas of applications; (3) identification of the themes and methodologies which have not been explored much in bioprocessing and chemical engineering applications, like the use of scientific knowledge to help improve the ML model architecture and learning process for more scientifically consistent solutions; and (4) illustrations of the use of these hybrid SGML methodologies applied to industrial polymer processes, such as inverse modeling and science-guided loss which have not been applied previously in such applications.

## 2 | APPLICATIONS OF HYBRID SGML APPROACH IN BIOPROCESSING AND CHEMICAL ENGINEERING

The integration of science-based models with data-based models has appeared in various fields like fluid mechanics[3], turbulence modeling[4], quantum physics[5], climate science[6], geology[7] and biological sciences.[8]

This study focuses on applications of hybrid SGML methodologies in bioprocessing and chemical engineering. Among the earliest applications is *the direct hybrid modeling* involving the integration of first-principles model with data-based neural networks[9]. Psichogios and Unger[10] combine a first-principles model based on prior process knowledge with a neural network, which serves as an estimator of unmeasured process parameters that are difficult to model from first principle. They apply the hybrid model to a fed-batch bioreactor, and the integrated model has better properties than the standard "black-box" neural network models. In particular, *the integrated model is able to interpolate and extrapolate much more accurately, is easier to analyze and interpret, and requires significantly fewer training examples*. Thompson and Kramer[11] later demonstrate how to integrate simple process model and first-principles equations to improve the neural network predictions of cell biomass and secondary metabolite in a fed-batch penicillin fermentation reactor when trained on sparse and noisy process data.

Agarwal[12] develops a general qualitative framework for identifying the possible ways of combining neural networks with the prior knowledge and experience embedded in the available first-principles models, and discusses *the direct hybrid modeling with series or parallel configuration* to combine the outputs of the science-based model and the ML model. Asprion, et al.[13] present the term, *grey-box modeling*, for optimization of chemical processes. They consider the case where a predictive model is missing for a process unit within a larger process flowsheet, and use measured operating data to set up hybrid models combining physical knowledge and process data. They report results of optimization using different gray-box models for process simulators applied to a cumene process. Actually, in a number of earlier studies, Bohlin and his coworkers have explored in details the concepts of gray-box identification for process control and optimization, and Bohlin has summarized the concepts, tools and applications of grey-box hybrid modeling in an excellent book.[14]

Over the years, we have seen a growing number of applications of hybrid modeling in bioprocessing and chemical engineering as part of the advances in smart manufacturing [15-17].

In their 2021 paper, Sansana et al.[16] discuss mechanistic modeling, data-based modeling, hybrid modeling structures, system identification methodologies, and applications. They classify their hybrid model into parallel, series, surrogate models (which are simpler mathematical representations of more complex models and similar to reduced-order models that we discuss below), and alternate structures (which include gray-box modeling mentioned above). In the alternate structures, they refer to some applications of semi-mechanistic model structures where the best hybrid model is selected using optimization concepts. They also classify the hybrid models based on some of the chemical industry applications into analysis of model-plant mismatch[17], model transfer, feasibility analysis and predictive maintenance, apart from the previous mentioned applications like process control, monitoring and optimization.

Von Stosch et. al.[18] have used the term, *hybrid semi-parametric modeling*, in their 2014 review, and have summarized applications in bioprocessing for monitoring, control, optimization, scale-up and model reduction. They emphasize that the application of hybrid semi-parametric techniques does not automatically lead to better results, but that rational knowledge integration has potential to significantly improve model-based process design and operation.

Qin and Chiang[19] review the advances in statistical machine leaning and process data analytics that can provide efficient tools in developing future hybrid models. In a latest paper, Qin et. al.[20] propose a statistical learning procedure integrating with process knowledge to handle a challenging problem of developing a predictive model for process impurity levels from more than 40 process variables in an industrial distillation system. Both studies highlight the power of statistical machine leaning for developing future hybrid process models.

A survey of the literature has shown applications of hybrid modeling in bioprocesses [21-27], chemical and oil and gas process industries[28-32], and polymer processes [33,34] for more accurate and scientifically consistent predictions. This survey has also shown many topical focuses of applications in bioprocessing and chemical engineering, including process control [35-38], design of experiments [39,40], process development and scale-up [41,42], process design[43] and optimization [13,44,45].

In a recent study, Zhou et al.[46] present *a hybrid approach for integrating material and process design* that holds much promise in process and product design. Cardillo et. al.[47] demonstrate the importance of hybrid models in silico production of vaccines to accelerate the manufacturing process. Chopda et. al.[23] apply integrated process analytical techniques, and modeling and control strategies to enable the continuous manufacturing of monoclonal antibodies. McBride et. al.[48] classify the hybrid modeling applications in different separation processes in chemical industry, namely, distillation [49-51], crystallization [52,53], extraction [54-56], floatation [57,58], filtration [59,60] and drying [61,61]. Venkatasubramanian[63] gives an excellent exposition of the current state of development and applications of artificial intelligence in chemical engineering. The author highlights the intellectual challenges and rewards for developing the conceptual frameworks for hybrid models, mechanism-based causal explanations, domain-specific knowledge discovery engines, and analytical theories of emergence, and presents examples from optimizing material design and process operations.

In an excellent edited volume, Glassey and Stosch[64] discuss some of the key strengths of hybrid modeling in chemical processes, particularly in the prediction of scientifically consistent results beyond the experimentally tested process conditions, which is crucial for process development, scale-up, control and optimization. They also identify some challenges. For example, incorrect fundamental knowledge in a science-based model could impose bias on predictions, thus the underlying assumptions used in a model are important for analysis. Also, time and accuracy of parameter estimation is critical when deciding on a hybrid modeling strategy. Kahrs and Marquardt[65] discuss the approach of simplifying the complex hybrid models into sequence of simpler problems, such as data preprocessing, solving nonlinear equations, parameter estimation and building empirical models using ML.

Herwing and Portner in their latest book showcase the applications of hybrid modeling in digital twins for smart biomanufacturing [124].

A recent patent by Chan et al.[66] presents Aspen Technology's approach on asset optimization using integrated modeling, optimization and artificial intelligence. In a later white paper, Beck and Munoz[67] describe Aspen Technology's current focus on hybrid modeling, combining AI and domain expertise to optimize assts. In particular, based on their application experience in in chemical industries, Aspen Tech have classified hybrid models into three categories: AI-driven, first-principles driven and reduced-order models [67]. They define *an AI-driven hybrid model* as an empirical model based on plant or experimental data and use first principles, constraints and domain knowledge to create a more accurate model. Examples of AI-driven models are inferential sensors or online equipment models. They define *a first-principles driven hybrid model* as an existing first-principles model augmented with data and AI to improve model's accuracy and predictability, which has seen many applications in bioprocessing and chemical engineering. Lastly, they define *a reduced-order model* where we use ML to create an empirical data-based model based on data from numerous first-principles process simulation runs, augmented with constraints and domain expertise, in order to build a fit-for-purpose low-dimensional model that can run more quickly. With reduced-order models, we can extend the scale of modeling from units to the plant-wide models that can be deployed faster.

# 3 | A CLASSIFICATION AND EXPOSITION OF HYBRID SCIENCE-GUIDED MACHINE LEARNING MODELS

As we have seen thus far, the majority of work in hybrid model applications in bioprocessing and chemical engineering focuses on the direct combination of science-based and data-based models. In this article, we portray a broad perspective of the combination of scientific knowledge and data analysis in bioprocessing and chemical engineering as inspired by some of the applications in physics and other areas [1,2]. We categorize these hybrid SGML applications in chemical process industry into two major categories, namely, ML compliments science and science compliments ML, together with their sub-categories based on the methodologies and objectives of hybrid modeling as illustrated in Figure 1. We also classify the applications in bioprocessing and chemical engineering according to our hybrid SGML approach. We present examples in several areas of SGML which have not been explored much thus far, and which have great potential for process improvement and optimization.

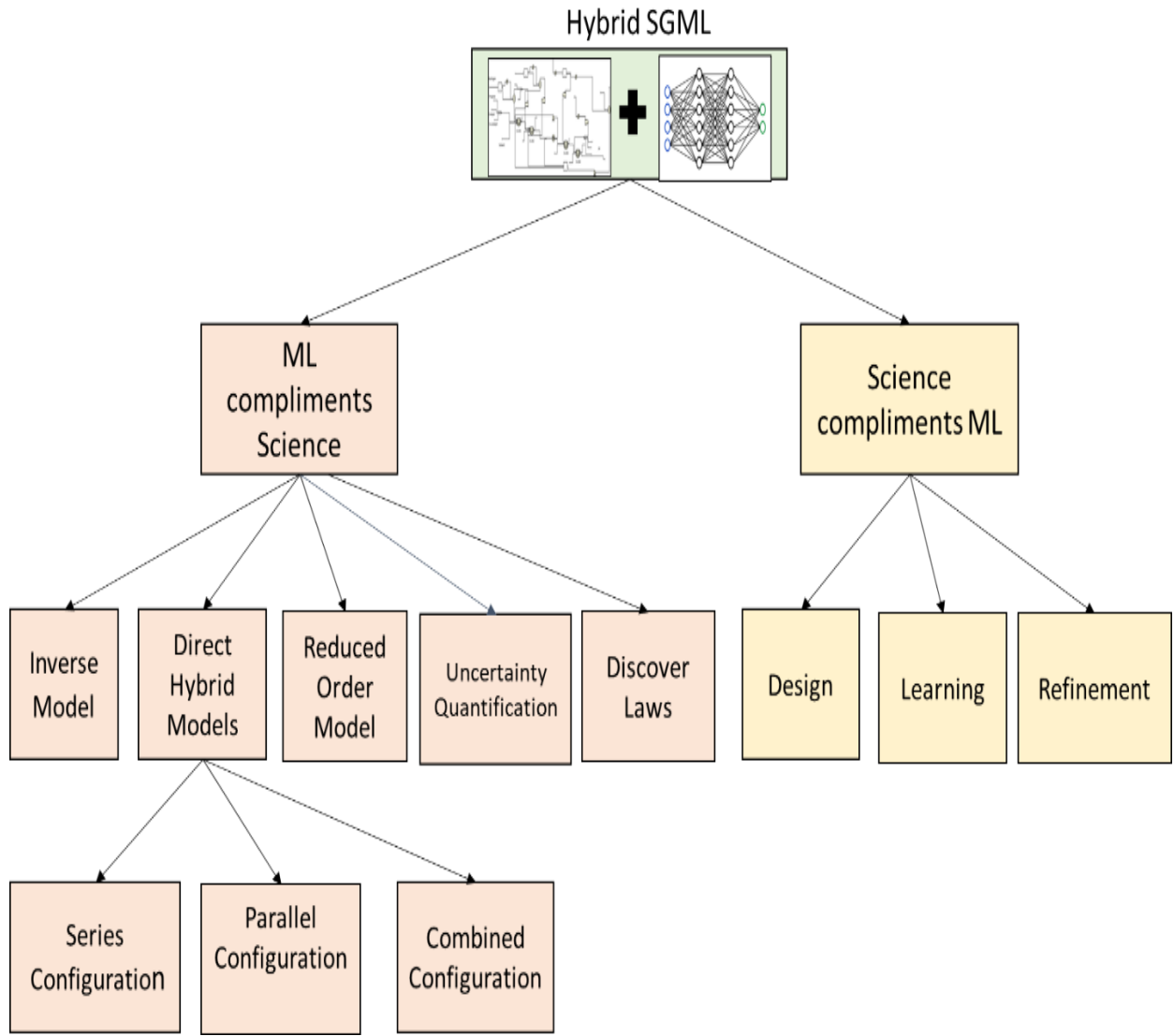

Figure 1. Classification of hybrid SGML models

## 4 | ML COMPLEMENTS SCIENCE

We can integrate a first-principles scientific model with a data-based model to improve the model accuracy and consistency. In the following, we introduce the sub-categories of direct hybrid modeling, inverse modeling approach, reducing model complexity, quantifying uncertainty in the process, and discovering governing equations.

### 4.1 | Direct Hybrid Modeling

A direct hybrid model combines the output of a first-principles or science-based model with the output of a data-based ML model to improve the prediction accuracy of dependent variables. These combinations could occur in a series configuration, a parallel configuration, or a series-parallel configuration. The direct hybrid modeling strategy is the most widely used approach in hybrid modeling in bioprocessing and chemical engineering.

#### 4.1.1 | Parallel Direct Hybrid Model

Figure 2 illustrates the concept of a parallel direct hybrid model. The science-based model may use the initial conditions and boundary conditions as inputs to make a prediction (Ym), while the ML model uses dynamic time-varying data to make the predictions (Yml). We then combine both outputs directly or with assigned weights (w1, w2) to achieve higher prediction accuracy. We can determine the weights by least squares optimization to minimize the total sum of squares of errors for the difference between the plant and the hybrid model.

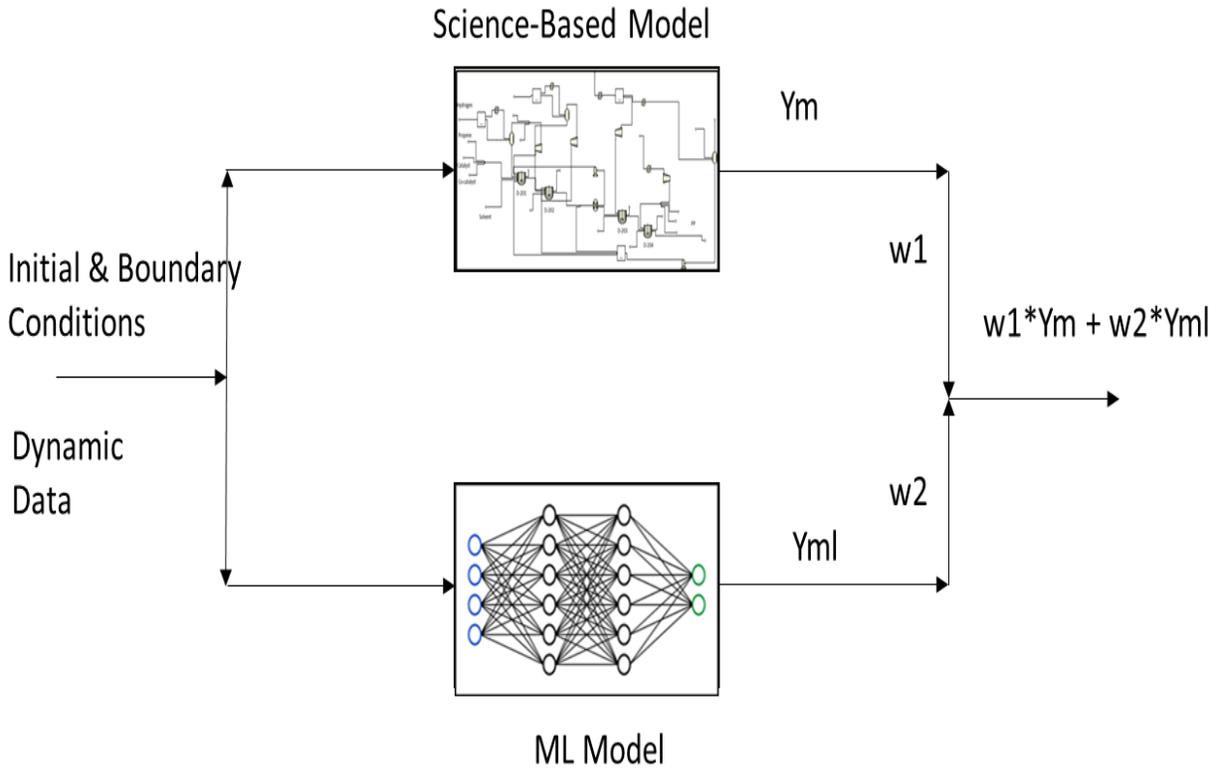

Figure 2. Parallel direct hybrid model: Ym and Yml are model predictions, and w1 and w2 are weights.

Galvanauskas et. al.[68] combine directly the data-based neural networks for kinetics and viscosity predictions with the first-principles mass balance ordinary differential equations to optimize the production rate of an industrial penicillin process. Chang et. al.[33] showcase a parallel hybrid model for the dynamic simulation of a batch free-radical polymerization of methyl methacrylate. They combine an approximate rate function for the concentration of the immeasurable initiator concentration with a black-box time-dependent or recurrent neural network model[9] of the dependent variables representing the mass and moment balance equations of the polymerization reactor. They use the resulting hybrid neural network and rate function (HNNRF) model to optimize the batch polymerization system, identifying the optimal recipe or operating conditions of the batch polymerization system.

*Hybrid residual modeling* or *parallel direct hybrid residual model* is a class of the parallel direct hybrid model, where we use a first-principles or science-based process model to quantify the time-dependent prediction error or residual, Yres, between plant data Y(t) and science-based model prediction Ym as a function of process variables [41,69-71]. Figure 3 illustrates the concept of the parallel direct hybrid residual model. The correction to the model output taking care of the prediction error or residual of the ML model in the hybrid residual configuration improves the model accuracy over the non-residual configuration of Figure 2.

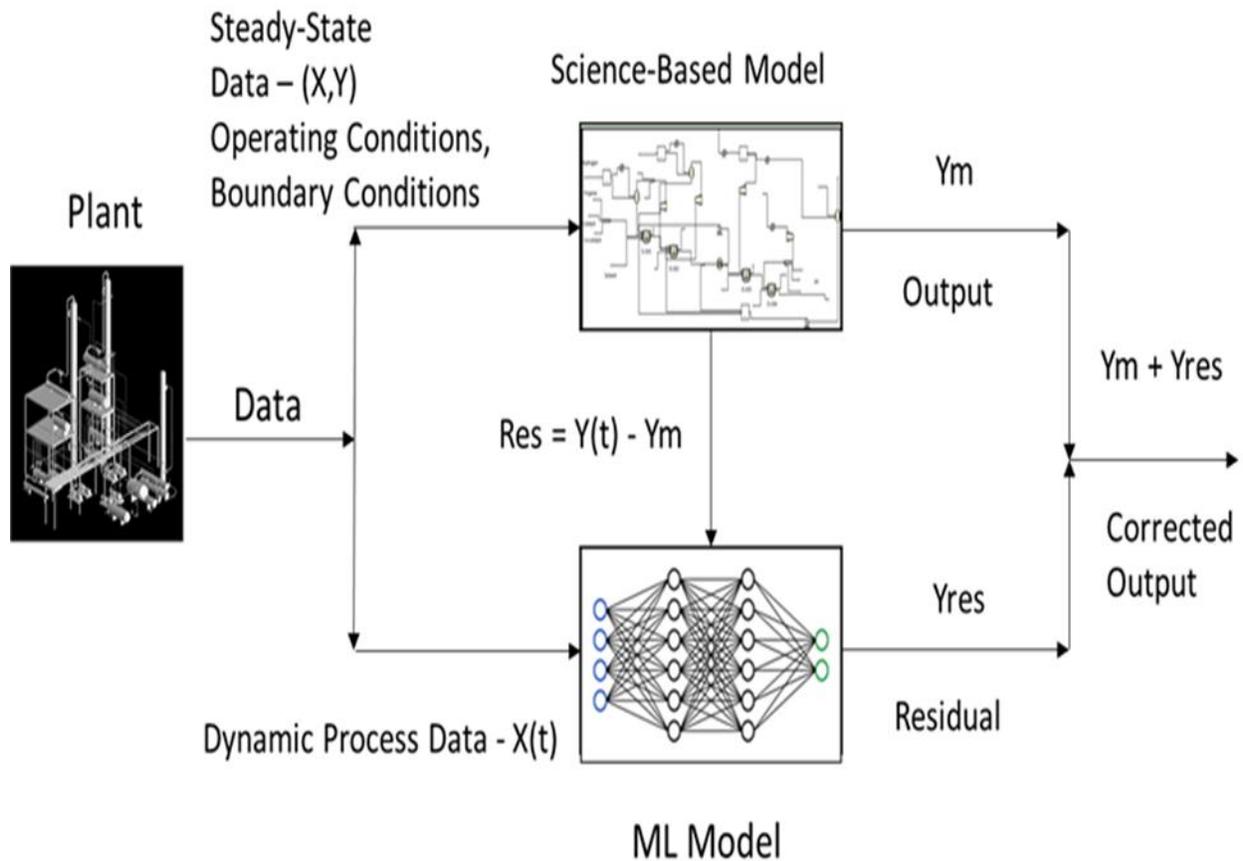

Figure 3. Parallel direct hybrid residual model: Ym represents model outputs, Res are the time-dependent prediction errors or residues between plant data Y(t) and science-based model outputs Ym, and Ym + Yres are the corrected model outputs

We recommend that the use of hybrid models will generally perform better than standalone ML model for applications like process development. This follows because hybrid models are better at extrapolation, while standalone ML models can be adequate for prediction in a steady running plant.

Tian et al.[69] develop a hybrid residual model for a batch polymerization reactor. First, they develop a simplified process model based on polymerization kinetics, and mass and energy balances to predict the monomer conversion, number-average molecular weight MWN, and weight-average molecular weight MWW. This first-principles process model cannot predict these product quality targets accurately because of its neglect of the gel effect at high monomer conversion and other factors. Next, the authors develop a parallel configuration of three data-based, time-dependent or recurrent neural networks[9] trained by process data to predict the residuals of monomer conversion, MWN and MWW of the simplified first-principles process model. The predicted residuals are added to the predictions from the simplified process model to form the final hybrid model predictions. Because of focus in batch process control is on the end-of-batch product quality targets, the use of time-dependent or recurrent neural networks can usually offer good long-range predictions. Therefore, the resulting hybrid residual model performs well in many batch process control and optimization applications [41,43,69-71].

Simutis and Lubnert[36] present another application of the direct hybrid modeling methodology to state estimation for bioprocess control. This work combines a first-principles state Kalman filter based on mass balances of biomass, substrate and product, and an ML-based observation model for quantifying relationship between less established variables and measurements. Recently, Ghosh et. al.[72-73] apply the parallel hybrid modeling framework in process control, where they combine first-principles models with data-based model built by applying subspace identification for better prediction of batch polymer manufacturing and seed crystallization system. Hanachi et. al.[74] showcase the application of direct hybrid modeling methodology for predictive maintenance. They combine a physics-based model with a data-based inferential model in an iterative parallel combination for predicting manufacturing tool wear.

**4.1.2 | Series Direct Hybrid Model**

Figure 4 illustrates the series direct hybrid model. The science-based process model serves to augment the data needs of the ML model, while the ML model can help in estimating the parameters of the science-based model. Babanezhad et al.[75] consider the computational fluid dynamics (CFD) for two-phase flows in chemical reactors, and couple science-based CFD results to a ML model based on an adaptive network-based fuzzy inference system (ANFIS). Once the ML model captures the pattern of the CFD results, they use the hybrid model for process simulation and optimization. Some features calculated from a science-based CFD model can *augment the data* as inputs to a ML model. Chan et. al.[66] have discussed the advantages of data augmentation by combining simulation and plant data to generate a more accurate data-based analysis. In an application to crude distillation in petroleum refining, Mahalec and Sanchez[51] use a science-based model to calculate the internal reflux to augment other plant data as inputs to a ML model, in order to calculate the relationship to the product true boiling point curves for quality analysis. The data augmentation in series hybrid models is more relevant when some feature measurements are missing in the original data, so we use a first-principles model to calculate those features and then augment those calculated data to the ML model to study the combined multivariate effects. The goal is more towards causal effect of the added science model features and less towards improving accuracy. If we find that some missing feature measurements cause a mismatch between a science-based model and the actual plant, data augmentation may improve the training performance of the hybrid model.

Krippl et. al.[76] present the hybrid modeling of an ultrafiltration process where they calculate the flux using a ML model to act as an input to a science-based model. Similarly, Luo et al.[29] develop a hybrid model for a fixed-bed reactor for ethylene oxidation, integrating first-principles reaction kinetics and reactor model with a ML catalyst deactivation model. The latter is developed with support vector regression from operating data, assuming the deactivation property decreasing monotonically with time. With the hybrid model, the prediction error is less than 5% for the prediction of an industrial reactor. The approach can predict the production more accurately and have more reliable extrapolation.

Figure 4 shows that a ML model can also help in *estimating the parameters* of the science-based model. Mantovanelli et al.[77] develop a hybrid model for an industrial alcoholic fermentation process, combining first-principles mass and energy balance equations for a series of five fermenters with a data-based, functional link network[75] to identify the kinetic parameters of the fermentation reactors trained by plant data. The hybrid model includes the effect of temperature on the fermentation kinetics and show good nonlinear approximation capability. Sharma and Liu[78] show how to use plant data to estimate kinetic parameters of first-principles models for industrial polyolefin processes. In a recent study Bangi and Kwong [125] estimate process parameters in hydraulic fracturing process using deep neural network which are then input to a first-principles model. Finally, we note that as illustrated in Figure 4, we can interchangeably use a science-based model or a ML model first in the hybrid framework, depending on if we require to add more features to augment the data set or to estimate model parameters.

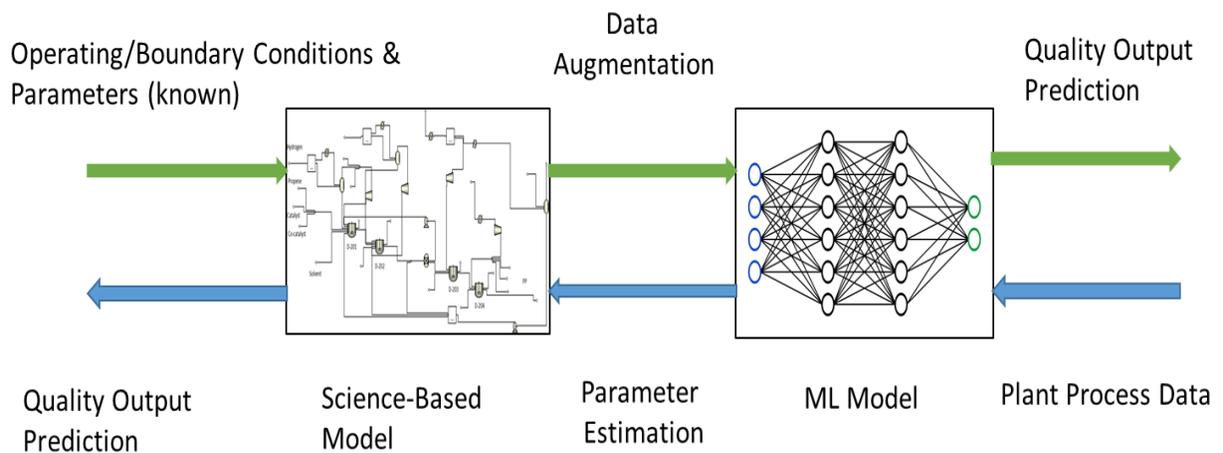

Figure 4. Series direct hybrid model

### 4.1.3 | Series-Parallel or Combined Direct Hybrid Model

Figure 5 shows a combined direct hybrid model, where we use the steady-state data from the plant to estimate the unknown parameters of a science-based process model and then uses the hybrid residual modeling strategy of Figure 3 for prediction. This series-parallel combination or feedback system can improve model predictions depending on the application.

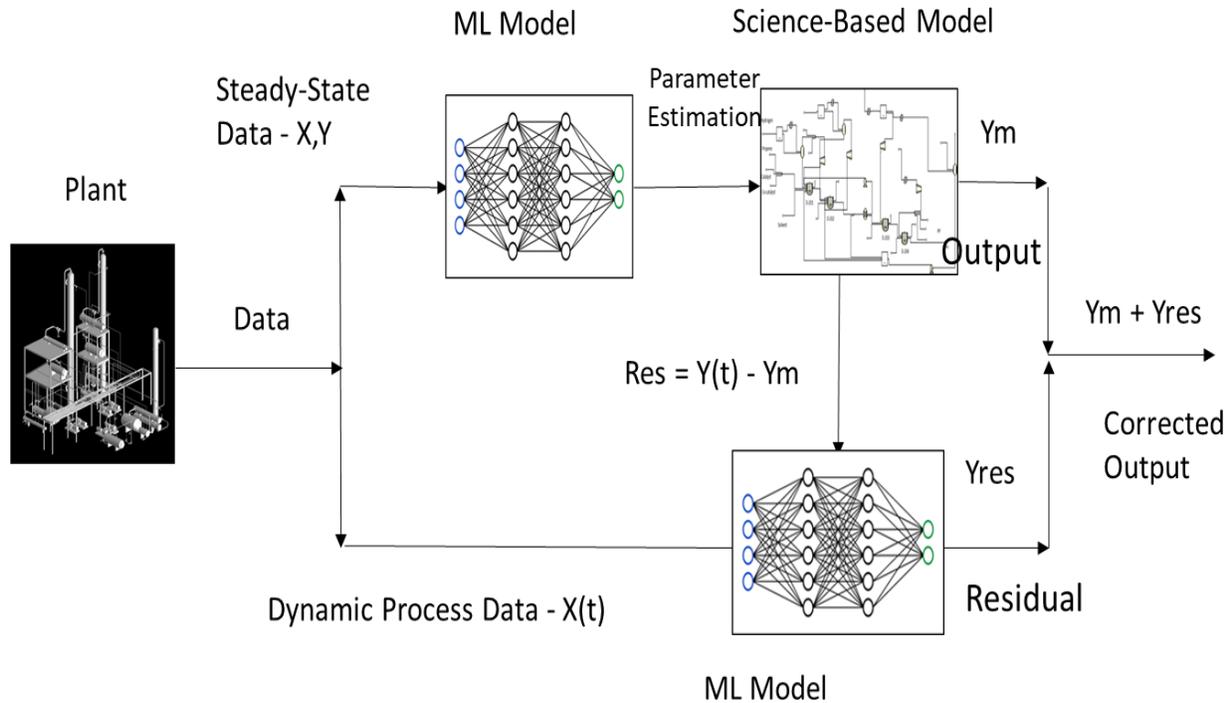

Figure 5. Combined Direct Hybrid Model: Ym are outputs, Yres are residuals, and Ym+ Tres are corrected outputs

Bhutani et. al.[79] present a definitive study comparing first-principles, data-based and hybrid models applied to an industrial hydrocracking process. In particular, they couple a first-principles hydrocracking model based on pseudocomponents with data-based neural network models of different configurations of Figures 3 to 5 that quantify the variations in operating conditions, feed quality and catalyst deactivation. The neural network component of the hybrid model either provides updated model parameters in the first-principles process model connected in series, or correct predictions of the first-principles process models. The hybrid models are able to represent the behavior of an industrial hydrocracking unit to provide accurate and consistent predictions in the presence of process variations and changing operating scenarios.

Song et. al.[80] also apply the direct hybrid model configurations of Figure 3 to 5 to an industrial hydrocracking process and analyze the strengths and weaknesses of these configurations. They call a model *a mechanism-dominated model* if the accuracy of its outputs is mainly dominated by the available theoretical knowledge used to develop the model; and they also call a model *a data-dominated model* if the accuracy of its outputs is mainly dominated by the quality of the training data and the performance of the resulting data-based model. In particular, they give both the first-principles model and the series direct hybrid model of Figure 4 as examples of mechanism-dominated models, and cite the data-based model, parallel direct residual model of Figure 3, and the combined direct hybrid model of Figure 5 as examples of data-dominated models.

In their work, Song et al.[80] combine a mechanism-dominated model with a data-dominated model as a hybrid direct model of Figure 2, with the weighting factors for the outputs of two individual models being determined in an adaptive fashion. For their application, Song et al. work with a mechanism-dominated model of an industrial hydrocracking process based on kinetic lumping[79,80], and with a data-dominated model based on a self-organizing map (SOM) followed by a convolutional neural network (CNN), with both being trained by simulated process data based on Aspen HYSYS[80]. They evaluate the performance of the hybrid model for operational optimization of the hydrocracking producing different product scenarios. While this study includes new conceptual development, it needs much simplification of its relatively complex methodology to make it readily applicable by data scientists and practicing engineers.

In a recent study, Chen and Lerapetritou[17] demonstrate how to use partial correlation analysis from multivariate statistics and mutual information analysis from information theory to identify and improve the plant-model mismatch in using a direct combined hybrid model for a pharmaceutical manufacturing process. As the authors state, implementing this plant-model mismatch strategy requires active excitation of variables online in order to capture the corresponding response data from the plant, which is often difficult to perform in manufacturing plants and in experimental settings, and could benefit from new development in computing and information technology.

Lima et al.[81] propose a semi-mechanistic model building framework based on selective and localized model extensions. They use a symbolic reformulation of a set of first-principles model equations in order to derive hybrid mechanistic–empirical models. The symbolic reformation permits the addition of empirical elements selectively and locally to the model. They apply the approach to the identification of a non-ideal reactor and to the optimization of the Otto–Williams benchmark reactor.

This combined strategy is generally more useful for the case where the science-based model has unknown parameters. We could use ML to determine these unknown parameters and then apply a hybrid residual ML approach. By doing so, we could improve the model prediction accuracy as well.

**4.1.4 | An Application of Combined Direct Hybrid Modeling to Polymer Manufacturing**

We apply the combined direct modeling strategy to an industrial polyethylene process for the prediction of melt index. We build a first-principles steady-state model of a Mitsui slurry high-density polyethylene (HDPE) process by following the methodology and kinetic parameters presented in Supplement 1b of Sharma and Liu[78]. For this application, it is easier to first estimate the complex multisite Ziegler-Natta polymerization kinetic parameters using steady-state production targets, and then convert the resulting steady-state simulation model based on Aspen Plus to a dynamic simulation model using Aspen Plus Dynamics. The resulting dynamic simulation model has similar independent process variables, including the feed flow and compositions and the reactor operating conditions. For less complex applications, dynamic data could be used for parameter estimation.

Figure 6 compares the predictions of the first-principles dynamic simulation model (in red) with the plant data with grade transitions (in green). We see much deviation between the model predictions and the plant data. We compare the MI values from the model with the plant data and calculate the error residuals. *The root-mean-squares-error (RMSE) values of the model residual is to 1.5* for the actual MI data with a standard deviation of 5.1.

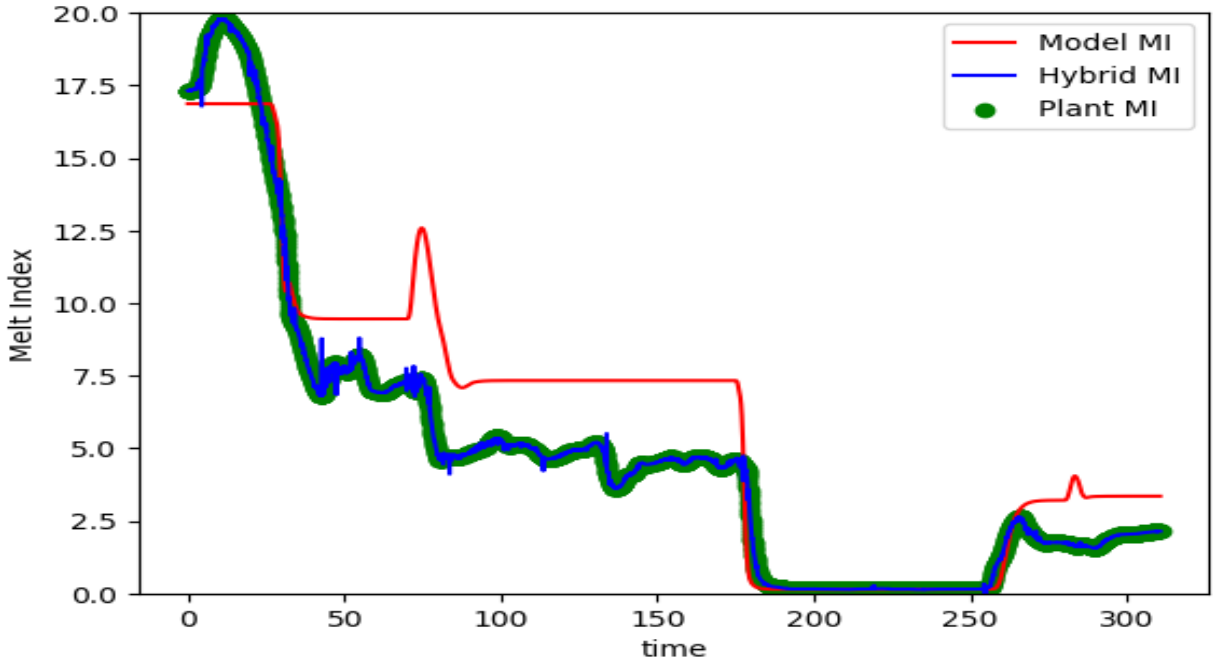

Figure 6.  Melt index prediction of a combined direct hybrid model compared to the first-principles model and plant data

To improve the accuracy of model predictions, we develop a regression model to predict the error residues as a function of independent process variables using a ML method called random forest algorithm[82] with Python. This leads to a hybrid model that predicts the MI value as a sum of the dynamic simulation model prediction (first-principles-based) and the predicted error residual (data-based) corresponding to a give set of independent process variable values, as illustrated in Figure 5. Figure 6 shows that the hybrid model predictions (*with a RMSE value of 0.21*) match the plant data much better than a first-principles dynamic simulation model alone. We note that a data-based model alone has also a similar accuracy, but it may give scientifically inconsistent results for predictions beyond process operating data which the model uses. Thus, the hybrid model is not only accurate, but also gives scientifically consistent results beyond current operating range.

**4.2 | Inverse Modeling**

In *inverse modeling*, we use the output of a system to infer its corresponding input or independent variables; this is different from the *forward modeling* where we use the known independent variables to predict the output of the system [2]. Figure 7 illustrates the inverse modeling framework. We see that in the traditional data-based approach, we use process variable data (X) and quality target data (Y) to train and test a ML model. Because the plant does not measure most quality targets continuously, we can apply a science-based process model, developed by first principles and validated by plant data, to predict and augment the quality target data (Y) for given process variable (X).

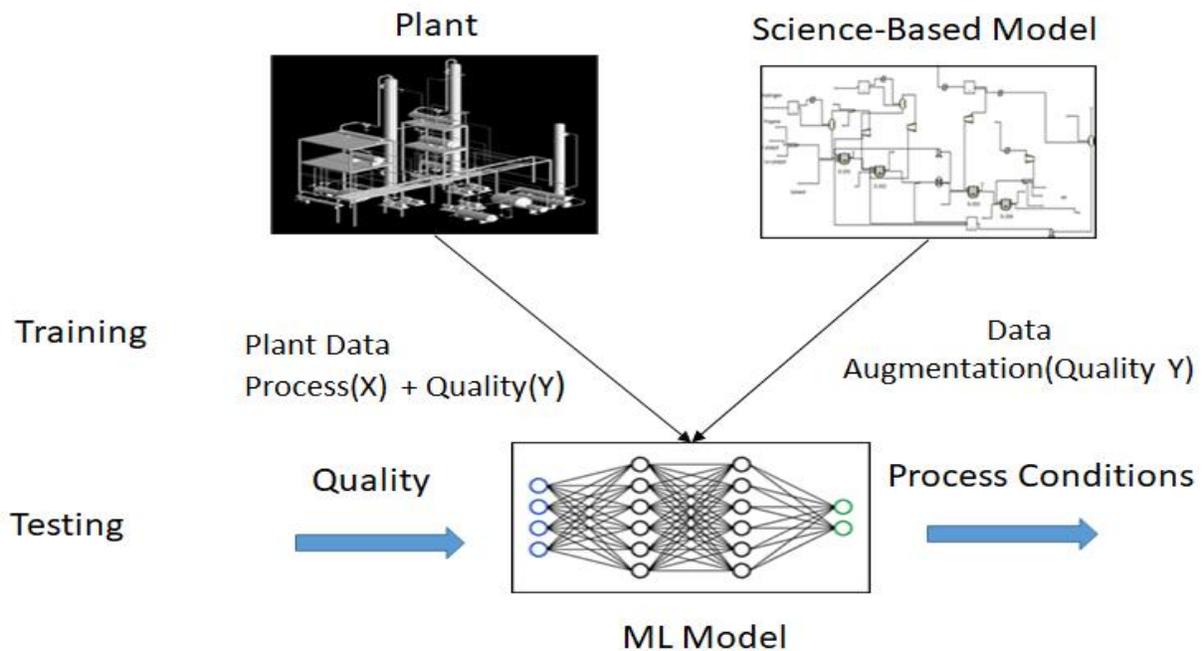

Figure 7. Inverse Modeling Framework

One of the earliest applications of inverse modeling for chemical process was by Savkovic-Stevanovic et. al.[83] They use a neural network controller for product composition control of a distillation plant based on the process inverse dynamic model relating the product composition to the reflux flow rate. The results illustrate the feasibility of using neural network for learning nonlinear dynamic model of the distillation column from plant input-output data. Their results also demonstrate the importance to take the time-delay of the plant into account.

Pharmaceutical product design and development typically uses the design of experiments (DOE) and response surface modeling (RSM) for steady-state process modeling, while neglecting the process dynamics and time delays. Tomba et. al.[84,85] demonstrate how to use the inverse modeling concept to generate process understanding with dynamic process models, quantifying the impact of temporal deviations and production dynamics. Specifically, they perform data-based, latent variable regression model inversion to find the best combination of raw materials and process variables to achieve the desired quality targets. The authors propose to combine design-of-experiments studies with hybrid modeling for process characterization.

Recently, Bayer al.[86] apply the inverse modeling approach to Escherichia coli fed-batch cultivations, evaluating the impact of three critical process variables. They compare the performance of a hybrid model to a pure data-driven model and the widely adopted RSM of the process endpoints, and show the superior behavior of the hybrid model compared to the pure black-box approaches for process characterization. The inverse modeling methodology makes the decision-making process in pharmaceutical product development faster, while minimizing the number of experiments and reducing the raw material consumption.

Raccuglia et.al.[87] train the ML learning model using reaction data to predict reaction outcomes for the crystallization of templated vanadium selenites. They demonstrate the use of ML to assist material discovery using data from previously unsuccessful or failed material synthesis experiments. The resulting ML model outperforms traditional human strategies, and successfully predicts conditions for new organically templated, inorganic product formation with a success rate of nearly 90%. Significantly, they show that inverting the machine-learning model reveals new hypotheses regarding the conditions for successful product formation.

There is a growing interest in the inverse approach to material deign, in which the desired target properties are used as input to identify the atomic identity, composition and structure (ACS) that exhibit such properties. Liao et al.[88] present a metaheuristic approach to material design that incorporates the inverse modeling framework.

Vankatasunramanian[61] also mentions the importance of inverse problem being solved by the application of artificial intelligence in chemical engineering processes.

Note the inverse modeling approach may lead to non-unique solutions which can give a range of predictions of input parameters within the operating range. By adding additional constraints to the input parameters (such as their operating range), we may obtain a unique solution.

**4.3 | An Application of Inverse Modeling to Polymer Manufacturing**

We illustrate the application of an inverse modeling approach that integrates steady-state and dynamic simulation models of a Mitsui slurry HDPE process, developed from first principles and validated by plant data, with a data-based ML model. The goal is to predict the operating conditions for producing new polymer grades, given the desired product quality targets, such as melt index (MI), polymer density (Rho), polydispersity index (PDI) and polymer production rate (P). The details of the steady-state simulation model are available in Supplement 1b of reference 78.

We first estimate the polymerization kinetic parameters from plant production targets in a steady-state model using Aspen Polymers based on our reported methodology[78]. This results in a validated Aspen Polymers steady-state simulation model. Next, we convert the steady-state model to a dynamic model using Aspen Plus Dynamics. We use the dynamic model to simulate the product quality data for different process operating conditions, which include the data characterizing the polymer grade transitions. Then, we use a Python-based, ensemble machine learning regression model[89] to regress the simulated data, with the simulated product quality data as input, and the process operating conditions (flow rates of all input streams) as the output. Given the desired quality targets for a new polymer grade, we apply the trained ML model to predict the operating conditions for the new polymer grade.

Figure 8 illustrates that the inverse modeling approach predicts the hydrogen feed flow rate with a high accuracy (low RMSE = 0.9) when compared to actual plant data for a standard deviation of 20). Thus, if we want to produce a new polymer grade given its quality targets, we can predict the operating conditions required to produce that polymer grade using the inverse modeling approach.

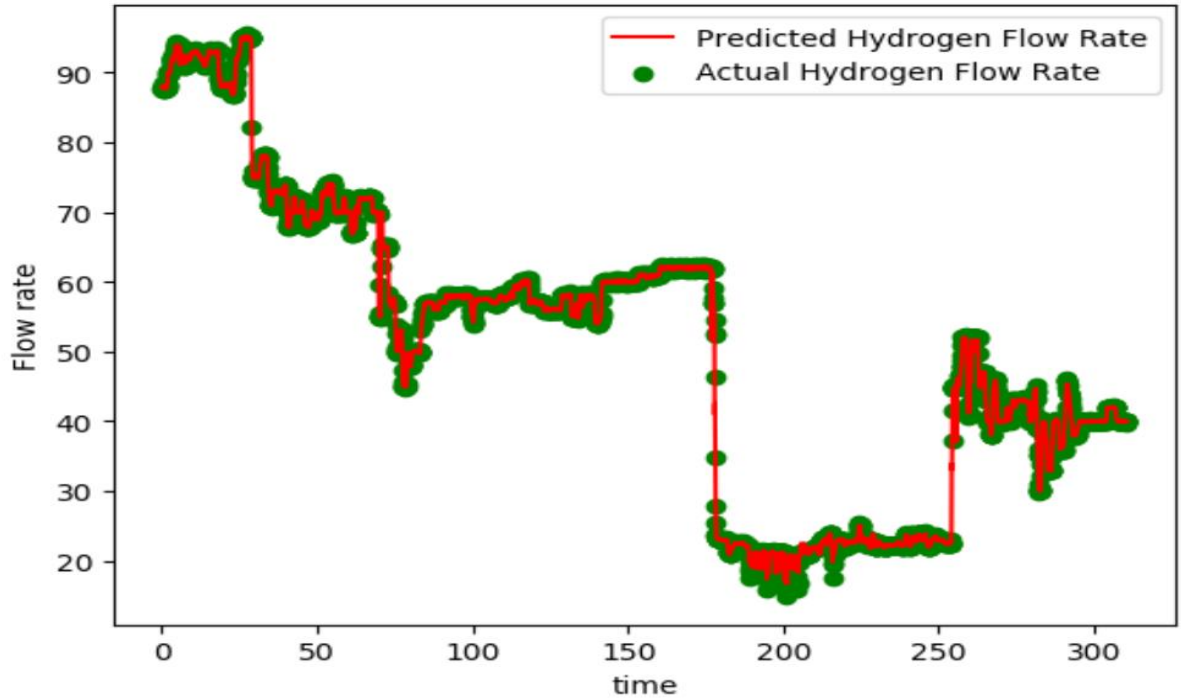

Figure 8. Hydrogen feed predictions from inverse modeling of product quality features

### 4.4 | Reduced-Order Models

Reduced-order models (ROMs) are simplified models that represent a complex process in a computationally inexpensive manner, but also maintain high degree of accuracy of prediction in simulating the process. In bioprocessing and chemical engineering, we can apply the ROM methodology to simulate complex processes and then use ML models to optimize the processes. See Figure 9. We can use ROMs to simulate different scenarios and sensitivities in order to generate process data, which in turn can be combined with ML models to build accurate soft sensors to predict quality variables. This approach helps to make sure that the ML model is trained on process data with multiple variations which is not possible in a steady plant run.

Hence, data-based sensors will be accurate for any future process optimization, scale up etc. and it is also easier to deploy such models online.

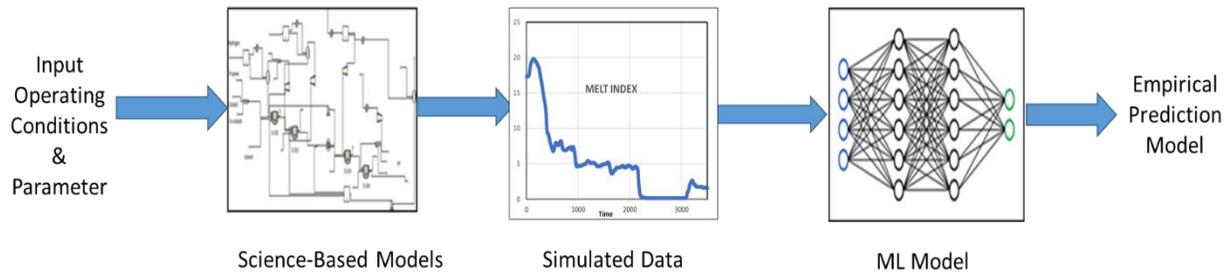

Figure 9. Reduced-order process modeling framework

In one of the earliest applications of ROM, MacGregor et. al.[90] apply a PLS (projection to latent squares or partial least squares) ML model of a polyethylene using process data simulated from a process model to develop inferential prediction models for polymer properties. This application involves a high-pressure tabular reactor system producing low-density polyethylene, in which all the fundamental polymer properties are extremely difficult to measure and are usually unavailable, and some on-line measurements such as the temperature profile down the reactor and the solvent flow rate are available on a frequent basis. The dimensionality reduction aspects of PLS facilitates the development of a multivariate statistical control plot for monitoring the operating performance of the reactors.

Model reduction can be achieved through dimensional reduction methods like principal component analysis. Another approach is to apply the residual combination with ML model for a ROM model, or to build a ML-based surrogate model for the full- order model. Reduced-order models have been called *surrogate models* in the context of grey-box modeling techniques where first-principles models are combined with data-based optimization techniques. Rogers and Lerapetritou[91,92] propose the use of surrogate models as reduced-order models that approximate the feasibility function for a process in order to evaluate the flexibility and operability of a science-based process model, since it is difficult to directly evaluate the feasibility due to black-box constraints.

In a recent study, Abdullah et. al.[93] showcase a data-based reduced-order modeling of non-linear processes that have time-scale multiplicity to identify the slow process state variables that can be used in a dynamic model. Agarwal et. al.[94] use ROM for modeling pressure swing adsorption process where they use a low-dimensional approximation of a dynamic partial differential equation model, which is more computationally efficient. In another study, Kumar et. al. [45] use a reduced-order steam methane reformer model to optimize furnace temperature distribution. In a recent study, Shafer et al.[95] use a reduced-dimensional dynamic model for the optimal control of air separation unit. The model combines compartmentalization to reduce the number of differential equations with artificial neural networks to quantify the nonlinear input–output relations within compartments. This wok reduces the size of the differential equation system by 90%, while limiting the additional error in product purities to below 1 ppm compared to a full-order stage-by-stage model.

Kumari et. al. [126] use data based reduced order methods for computational fluid dynamic modeling applied to a case study of super critical carbon dioxide rare event. They propose a k-nearest neighbor (kNN)-based parametric reduced-order model (PROM) for consequence estimation of rare events to enhance numerical robustness with respect to parameter change. Recently, many operator-theoretic modeling identification and model reduction approaches like the Koopman operators have been applied to integrate first-principles knowledge into finding relationship among multiple process variables in chemical processes. Koopman operator offers great utility in data-driven analysis and control of nonlinear and high-dimensional systems. Narsingham and Kwon [127] develop a new local Dynamic Mode Decomposition (DMD) method to better capture local dynamics which does temporal clustering of snapshot data using mixed integer nonlinear programming. The developed models are subsequently used to compute approximate solutions to the original high-dimensional system and to design a feedback control system of hydraulic fracturing processes for the computation of optimal pumping schedules.

Our focus on ROM is more towards using the science-based model to simulate process data that can be used by ML models to derive empirical correlations for process optimization. ROM are particularly useful in chemical processes for dynamic optimization of a complex large-scale process.

**4.5 | An Application of Reduced-Order Modeling to Polymer Manufacturing**

We illustrate the ROM methodology in a HYPOL polypropylene production process. The details of the steady-state simulation model are available in Supplement 1a of reference 78.

The Hypol process is complex with series of reactors, separators and recycle loops. The process has many operating variables, such as feed flow rates of propylene, hydrogen to each reactor, and temperature and pressure in each reactor. It is critical to quantify the effects of operating variables on the polymer quality targets, particularly melt index, in order to design or optimize the process. To achieve this, we need multivariate process data which are not usually available in a steady running plant. Hence, we use the ROM methodology.

We model the HYPOL polypropylene production process following the methodology of Sharma and Liu[78] and then run multiple steady-state simulations to generate multivariate data with varying operating variables and the corresponding melt index predictions. We use a random forest ML model[89] to train the simulated data to predict the melt index as a function of the process variables and also understand the causality of important features affecting the polymer quality. The empirical ML model can serve as an approximate quality sensor. We ca use it to predict the melt index at varying process variables. See Figure 10a.

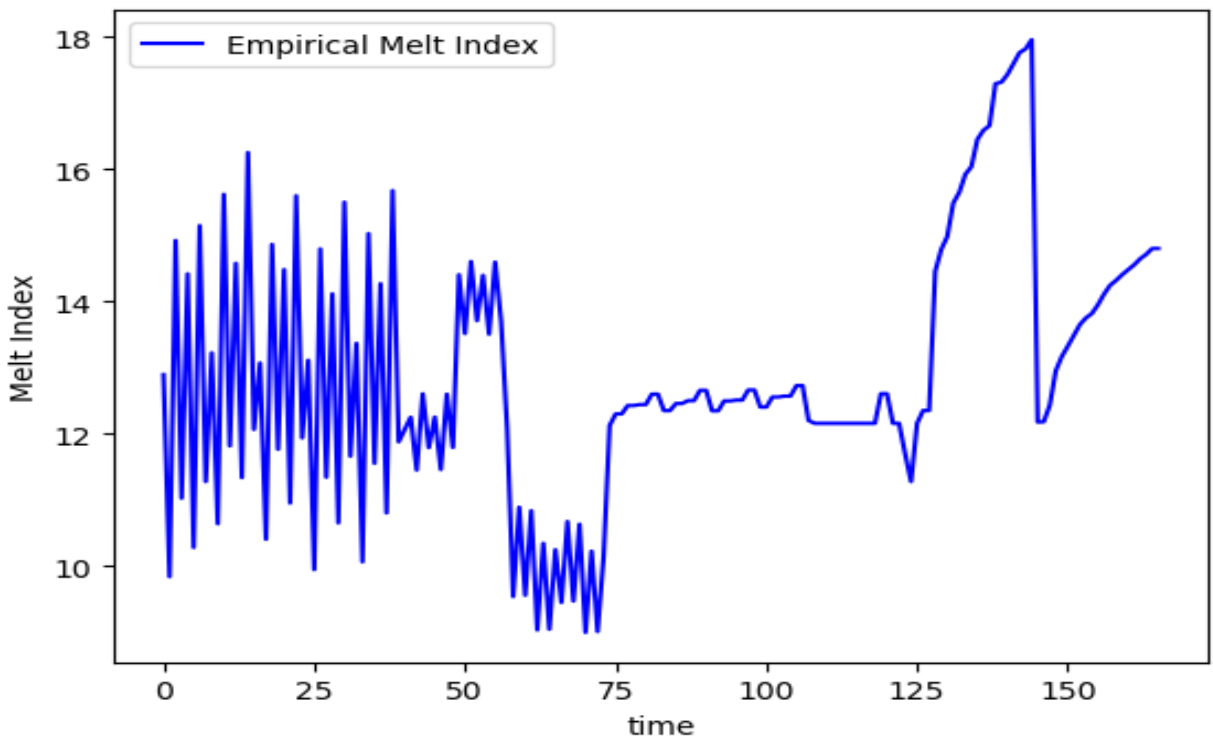

Figure 10. (a) Melt Index prediction at varying process variables

The ML model also decides the relative importance of different operating variables in reducing the mean decrease in "node impurity", which is a measure of how much each operating variable feature reduces the variance in the model. Figure 10b illustrates that the ROM calculates the important features like hydrogen flow rate (H24) and the temperature to the fourth reactor (R4T) as the most important variables affecting the melt index, which can then be used to find the optimum conditions to produce polymer of a specified melt index value and to improve the process design for a new process.

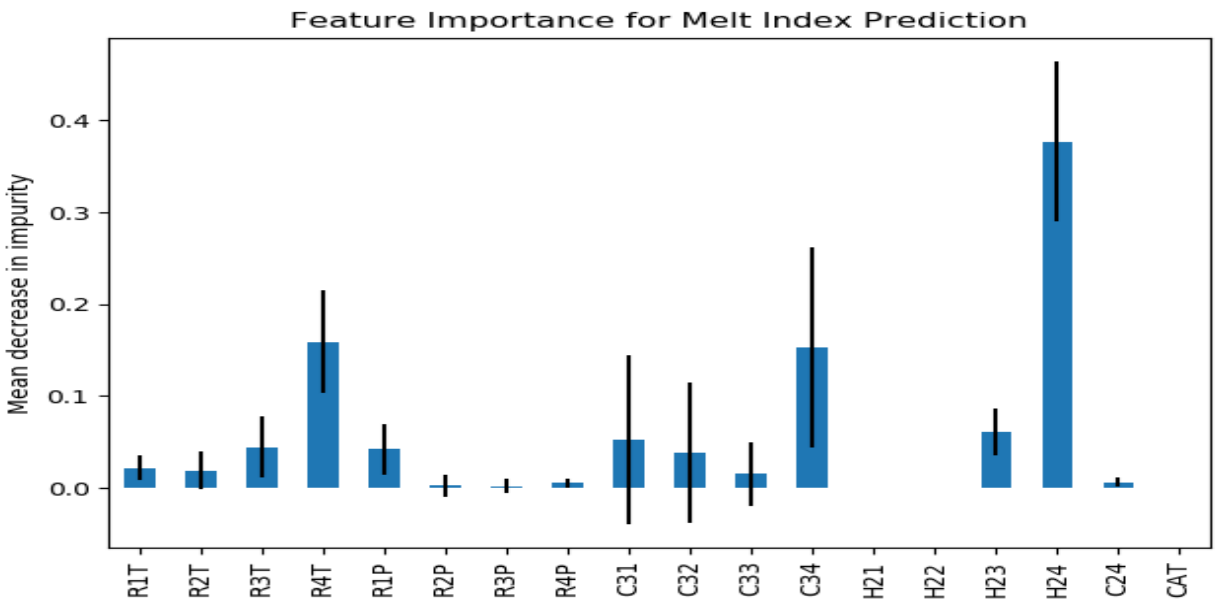

Figure 10. (b) Feature importance for melt Index prediction: RxT and RxP refer to the temperature and pressure of reactor x; C3x and H2x represent the mass flow rates of propylene and hydrogen to reactor x; C24 is the mass flor rate of ethylene to reactor 4; and CAT is the catalysts mass flow rate.

**4.6 | Hybrid SGML Modeling for Uncertainty Quantification**

A science-based model can produce results with some uncertainties which can be quantified by ML-based techniques. The uncertainties in science-based models arise from uncertainty in model parameters, and boundary and initial conditions. In some cases, the model bias and assumptions can be a source of uncertainty as well. We can use the predictions from a calibrated model to quantify uncertainties. Data-based ML models like Gaussian process, neural networks etc. are used to help build a surrogate model that defines a relation between model inputs and outputs which can then be used to quantify the uncertainty.

Because of uncertainty in process inputs and process states in a chemical process model, the uncertainty propagates to the process outputs as well. The uncertainty in a science-based model due to any of the parameters or any of the prior knowledge can be used by a ML model to quantify uncertainty in a chemical process as shown in Figure 11.

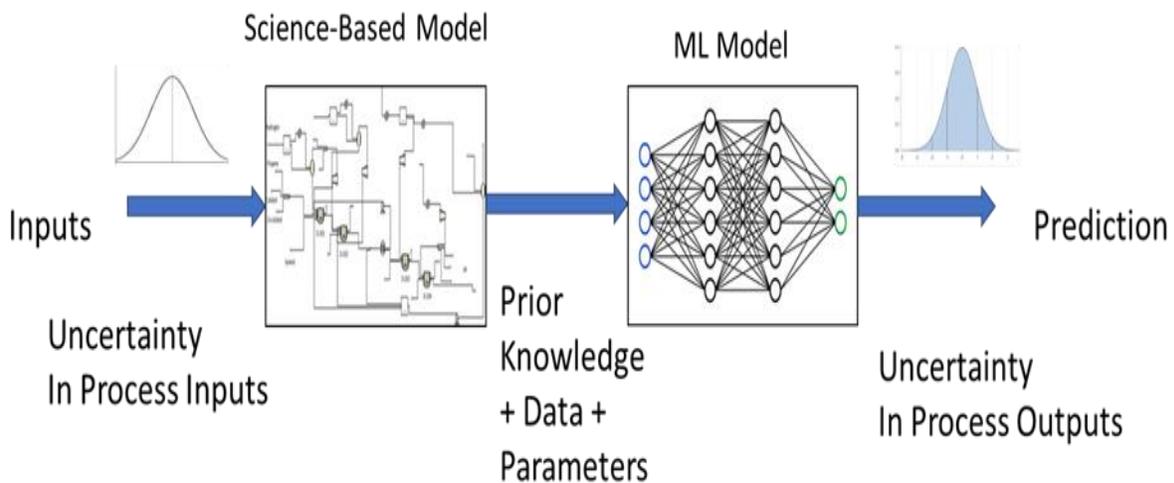

Figure 11. Uncertainty quantification of melt Index prediction of a slurry HDPE process

This surrogate data- based ML modeling reduces the computational expense of Monte Carlo methods, which are traditionally used for uncertainty quantification (UQ)[96].

Because of uncertainty in process inputs and process states in a chemical process model, the uncertainty propagates to the process outputs as well. Duong et. al.[97] uses UQ for process design and sensitivity analysis of complex chemical processes using the polynomial chaos theory. Fenila et. al.[98] utilize UQ for electrochemical synthesis, where they calculate simulation uncertainties and global parameter sensitivities for the hybrid model. UQ has also been applied to understand complex reaction mechanisms. Proppe et. al.[99] showcase kinetic simulations in discrete-time space considering the uncertainty in free energy and detecting regions of uncertainty in reaction networks. UQ techniques are popular in the field of catalysis and material science as they are used to quantify the uncertainty of models based on density functional theory[100,101]. In another study, Boukouval and Lerapetritou[102] demonstrate the feasibility analysis of a science-based process model over a multivariate factor space. They use a stochastic data-based model for feasibility evaluation, referred to as Kriging and develop an adaptive sampling strategy to minimize sampling cost while maintaining feasibility.

## 4.7 | An Application of SGML Modeling to Uncertainty Quantification in Polymer Manufacturing

We quantify the uncertainty of the chemical process model in predicting the melt index for the industrial HDPE process described in Section 2.1.4. This uncertainty in prediction may result from the estimated kinetic parameters of the process, which propagates to the quality output as well. We simulate the data using the chemical process model and calculate the prediction intervals using a gradient boosting ML model[89]. In this case, we use the concept of prediction intervals to determine the range of model prediction. We use the quantile regression loss with gradient boosting model to predict the prediction intervals [103]. We define a lower and an upper quantile according to the desired prediction interval. Figure 12 illustrates the uncertainty in the prediction of melt index given by the range of the 90% prediction interval which implies that there is 90% likelihood that the ML model prediction will lie in the given range. The resulting RMSE value lies within 1.2 to 1.5, with the standard deviation of melt index data equals 5.1. In the figure, we see that the prediction interval is the area between the two black lines represented by the upper quantile (95$^{th}$ percentile) and the lower quantile (5$^{th}$ percentile). From the figure, we see a larger prediction interval that means a higher uncertainty in prediction for time less than 100 hours compared to the later stage because of a more appreciable change in MI in that interval. Thus, uncertainty quantification (UQ) helps in making better process decisions after knowing the error estimate of the model.

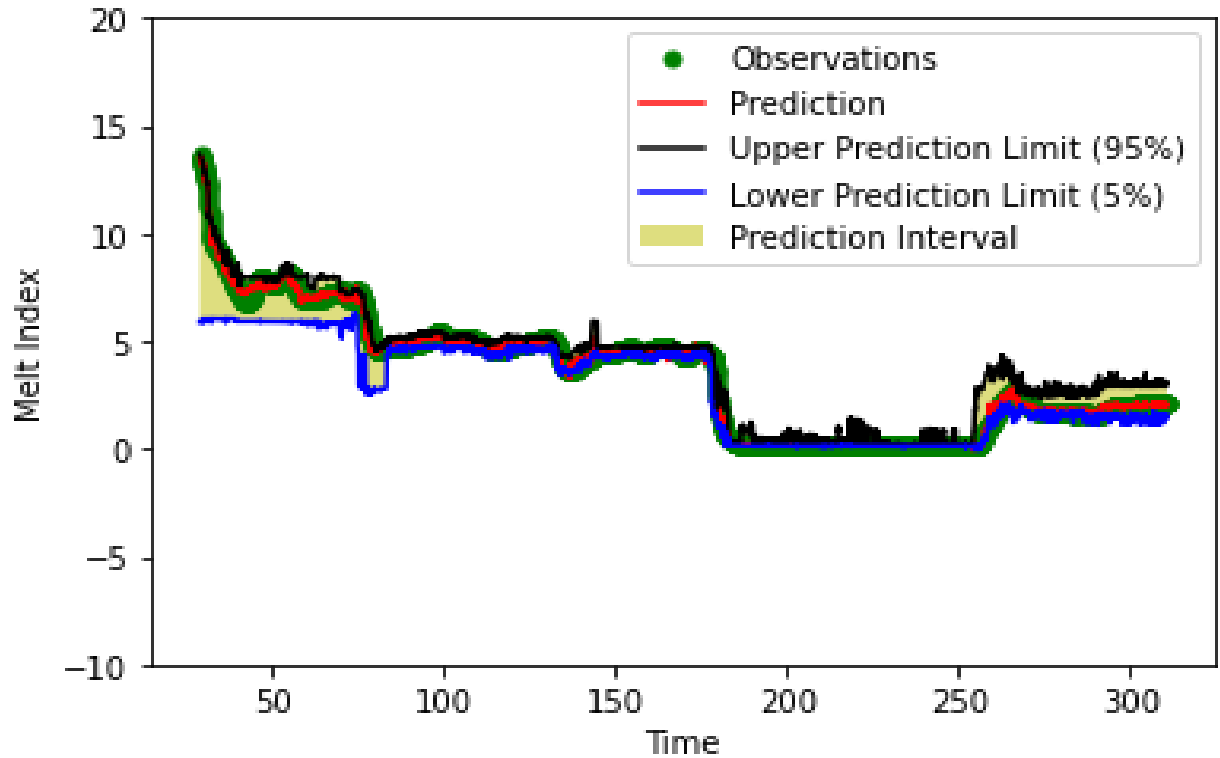

Figure 12. Uncertainty quantification of melt Index prediction of a slurry HDPE process

**4.8 | Hybrid SGML Modeling to Aid in Discovering Scientific Laws Using ML**

One way in which ML can help science-based modeling is by discovering new scientific laws which governs the system. There is a growing application of ML in physics to rediscover or discover physical laws mainly by data-driven discovery of partial differential equations. ML can be used to develop an empirical correlation which can be used as a scientific law in a science-based model, or ML can be used to solve the partial differential equation defining scientific laws as illustrated in Figure 13.

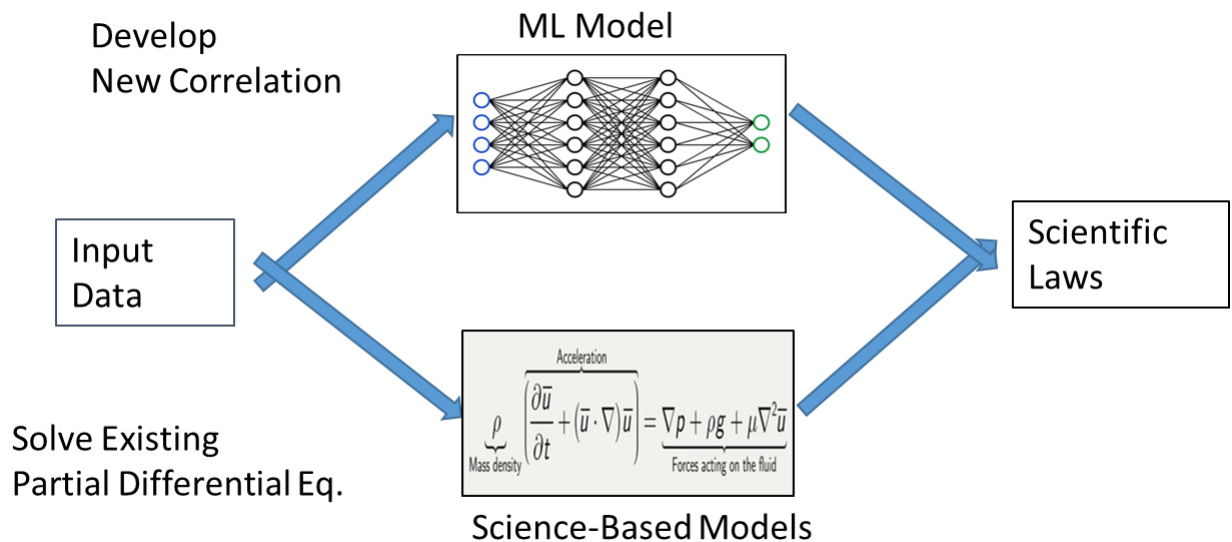

Figure 13. Discovering scientific laws

Rudy et. al.[104] showcase the discovery of physical laws like the Navier-Stokes equation and the reaction-diffusion equation in chemical processes by a sparse regression method governing the PDE by using a system of time series measurements. Langley et. al.[105] present the applications of ML in rediscovering some of the chemistry laws, such as the law of definite proportions, law of combining volumes, determination of atomic weights and many others.

Another important application of ML is to discover some of the thermodynamic laws which can be useful in defining the phase equilibrium and critical for an accurate science-based process model. Nentwich et. al.[106] use data-based mixed adaptive sampling strategy to calculate the phase composition, instead of the complex equation-of-state models. Thus, ML application can have promising use in discovering more accurate physical and chemistry laws that govern the chemical process. This methodology can be used to obtain the functional form of scientific laws as well as the estimation of the parameters of existing laws. Brunton et al.[128] demonstrate a novel framework to discovering governing equations underlying a dynamic system simply from data measurements, leveraging advances in sparsity techniques and machine learning. These scientific laws calculated by ML-based models can then be utilized in first-principles model to improve accuracy as well as reduce model complexity.

## 5 | SCIENCE COMPLIMENTS ML

Referring to Figure 1, we can also improve ML models using scientific knowledge. We can improve *the generalization or extrapolation capability* and reduce the scientific inconsistency of ML models by using scientific knowledge in designing the ML models. The scientific knowledge can also help in improving the architecture of the data-based ML model or the learning process of the ML model and even with the final post-processing of the ML model results.

### 5.1 | Science-Guided Design

In science-guided design, we choose the model architecture based on scientific knowledge. For a neural network, we can decide the intermediate variables expressed as hidden layers based on scientific knowledge of the system. This helps in improving the interpretative ability of the models. Figure 14 illustrates a neural network model whose architecture like the number of neurons, hidden layers, activation layers etc. can be decided by prior scientific knowledge.

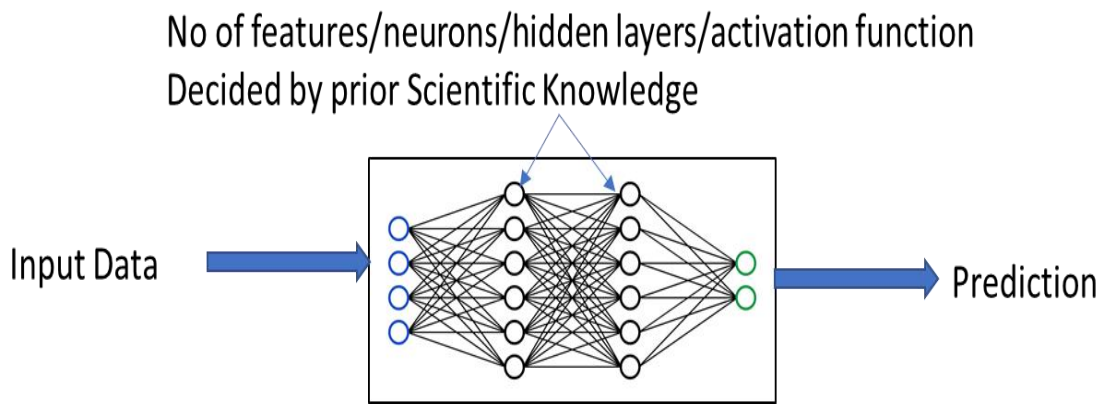

Figure 14. Science-guided design framework of neural network architecture

In a bioprocess application, Rodriguez-Granrose et. al.[39] use the design of experiments (DOE) to create and evaluate a neural network architecture. They use DOE to evaluate activation functions and neurons on each layer to optimize the neural network. In their recent study, Wang et al.[107] design their theory-infused neural networks based on adsorption energy principles for interpretable reactivity prediction. The use of the novel neural differential equation[108] to solve a first-principles dynamic system represents a hybrid SGML approach, where the architecture of ML model is influenced by the system and finds applications in continuous time series models and scalable normalizing flows. The derivative of hidden state is parameterized using a neural network and the output of the network is computed using a differential equation solver. In a recent study, Jaegher et. al.[109] use the neural differential equation to predict the dynamic behavior of electro-dialysis fouling under varying process conditions. In a recent application of this theme in chemical process for model predictive control, Wu et. al.[110] use prior process knowledge to design the recurrent neural network (RNN) structure[9].They showcase a methodology to design the RNN structure using prior scientific knowledge of the system and also employ weight constraints in the optimization problem of the RNN training process. Reis et. al.[111] discuss the concept of incorporation of process-specific structure to improve process fault detection and diagnosis.

Fuzzy artificial neural networks (ANN) is a class of neural networks which utilize prior scientific knowledge of the system to formulate rules mapped on to the structure of the ANN[9,112]. The weights of the ANN connecting the process input to output can be connected to physical process variables [64]. Apart from making the models more scientifically consistent with prior knowledge, they also reduce computational complexity and provides interpretable results. The use of prior knowledge also makes them suitable for extrapolation. Fuzzy ANN have been particularly useful for applications in process control [113]. Simutis et. al. use fuzzy ANN system for industrial bioprocess monitoring and control [114-115]. They also illustrate the application of fuzzy ANN process control expert to perform appropriate control actions based on process trends for bioprocess optimization and control [116].

Sparse Identification of Nonlinear Dynamics (SINDy) is another data-based modeling method that utilizes scientific knowledge for improving the model performance with the algorithms [128]. Bhadiraju et. al. [129] have used the SINDy algorithm to identify the Non Linear Dynamics of a chemical process system(CSTR). They used sparse regression in combination with feature selection to identify accurate models in an adaptive model identification methodology which requires much less than data that current methods. In a similar study Bhadiraju et. al. [130] have a modified adaptive SINDy approach that is helpful in cases of plant model mismatch and does not require retraining and hence computationally less expensive.

## 5.2 | Science-Guided Learning

Here, we make use of the scientific principles to improve the scientific consistency of data-based models by modifying the machine learning process. We do this by modifying the loss function, constraints and even the initialization of ML models based on scientific laws. Specifically, in order to make the ML models physically consistent we make the loss function of neural network model incorporate physical constraints [2]. A loss function in ML measures how far an estimated value is from its true value. A loss function maps decisions to their associated costs. Loss functions are not fixed, they change depending on the task in hand and the goal to be met. We can define a loss function (based on the mean squared error, MSE) of the ML model ($Loss_M$) for regression to calculate the difference between the true value ($Y_{true}$) and the model predicted value ($Y_{pred}$). Likewise, we can define a loss function for a science-based model ($Loss_{SC}$), which is a function of the model predicted value (Y_pred) consistent with science-based loss. We include a weighting factor λ to express the relative importance of both loss terms. We write the overall loss function (Loss) as:

$$Loss = Loss_M(Y_{true} - Y_{pred}) + \lambda Loss_{SC}(Y_{pred}) \quad (1)$$

Figure 15 illustrates the concept of science-guided loss function.

A science-guided initialization helps in deriving an initial choice of parameters before a model is trained so that it improves model training and also prevents from reaching a local minimum, which is the concept of transfer learning. Thus, we can use the data from a science-based model to pre-train a ML model based on this concept of initialization [1,2,7]. This concept has been utilized in chemical process model in the form of process similarity and developing new process models through migration. In particular, Lu et. al. [117] introduce the concept of process similarity, and classify it into attribute-based and model-based similarities. They present a model migration strategy to develop a new process model by taking advantage of an existing base model, and process attribute information. Adapting existing process models can allow using fewer experiments for the development of a new process model, resulting in a saving of time, cost, and effort. They apply the concept to predict the melt-flow-length in injection molding and obtain satisfactory results.

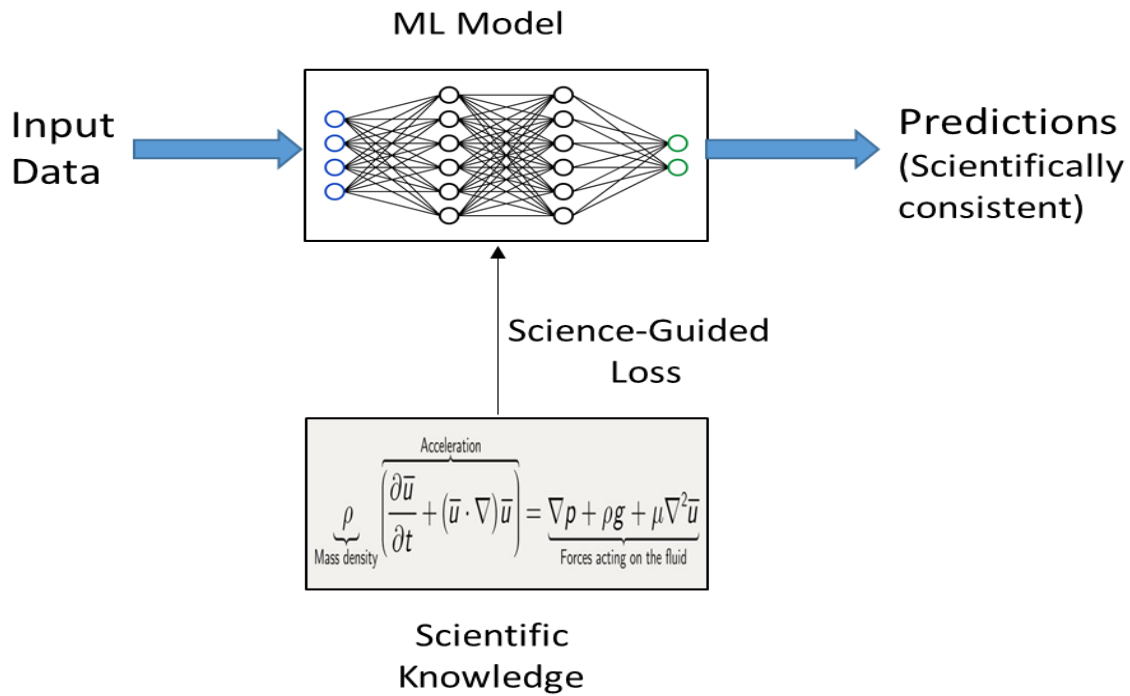

Figure 15. Science-guided loss function representation

In another study on the similar concept, Yan et al.[118] use a Bayesian method for migrating a ML Gaussian process regression model. They showcased an approach of an iterative model migration and process optimization for an epoxy catalytic reaction process.

Recently, Kumar et. al.[119] try to optimize the Non-Newtonian fluid flow for industrial processes like crude oil transportation using a physics- based loss function for the shear stress calculation for more accurate flow predictions. In another study on the similar principle, Pun et. al. [120] apply physics-informed neural networks for more accurate and transferable atomistic modeling of materials.

**5.3 | An illustrative Example of Science-Guided Learning**

We showcase the application of the science-guided loss function in the slurry HDPE process for the industrial HDPE process described in Section 2.1.4. The goal is to predict the melt index of the polymer. The plant only measures the polymer melt index as the quality output, but we also want the data-based ML model to predict the scientifically consistent polymer density values.

We express polymer density as a function of the melt index using some empirical correlations and modify the loss function (based on the mean squared error, MSE) to consider density as well. See Eq. (2) below. We then train a deep learning neural network model to predict the melt index of the polymer. Figure 16 illustrates that the SGML hybrid model calculates the melt index, resulting in a RMSE of the melt Index that is slightly higher (RMSE = 0.8) (standard deviation of data= 5) compared to a standalone ML model. In addition to predicting the melt index values, the hybrid SGML model is simultaneously predicting the polymer density correctly within the physically consistent range of 0.94-0.97 g/c. By contrast, the density estimates by the ML model alone result in density values greater than 1, which is physically inconsistent.

$$Loss = Loss_M(MI_{true} - MI_{pred}) + \lambda Loss_{SC}(\rho(MI_{true}) - \rho(MI_{pred})) \quad (2)$$

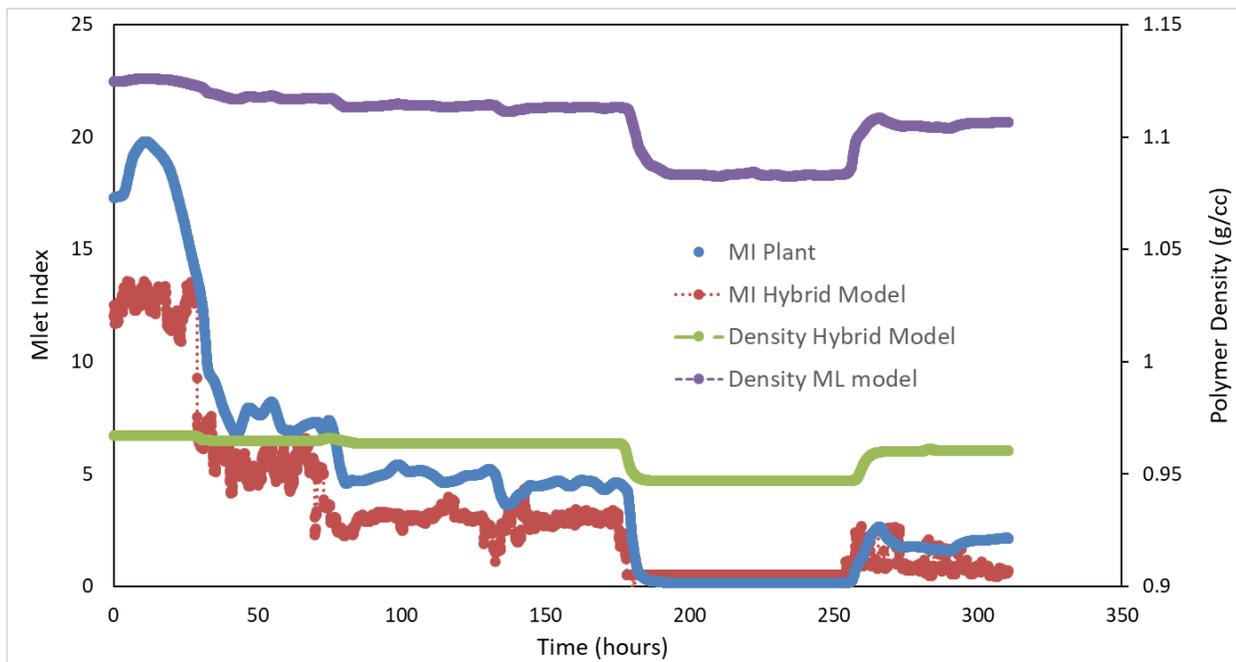

Figure 16. Melt index and polymer density prediction with a ML model with a science-guided loss function

**5.4 | Science-Guided Refinement**

By science-guided refinement, we mean the post-processing of ML model results based on scientific principles. This post-processing of results of the ML model using science-based models can be useful to the design and prediction of material structure[113]. Thus, the discovery of materials forms the basis of chemical process development from which the manufacturing process of any compound can be designed. This is different than the serial direct hybrid model discussed in Section 3.1.2. In particular, we use the science-based model to merely test the scientific consistency of the ML model results. Hautier et. al.[114] use first-s models based on density functional model to refine the results of probabilistic ML models to discovery ternary oxides. Figure 17 illustrates the science-guided refinement framework.

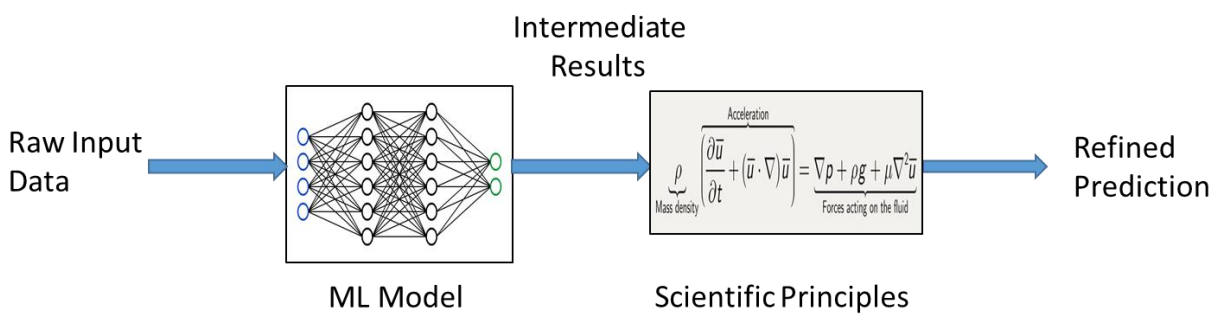

Figure 17. Science-guided refinement framework

Another application for science-guided learning is for data generation. ML techniques like generalized adversarial networks (GAN) are useful for generating data in an unsupervised learning. GANs do have a problem of high sample complexity[2] which can be reduced by incorporating some science-based constraints and prior knowledge. Cang et al. [115] apply ML models to predict the structure and properties of materials and use the results of the ab initio calculations to refine the ML model results. They generate more imaging data for property prediction using a convolution neural network and introduce a morphology constraint form scientific principles, while training of the generative models so that it improves the prediction of the structure- property model.

Thus, some of these methodologies of having science complimenting ML have much potential for future applications to bioprocessing and chemical engineering.

# 6 | CHALLENGES AND OPPORTUNITIES OF HYBRID SGML APPROACH FOR MODELING CHEMCAL PROCESSES

Along with all the merits of using the SGML methodology there are challenges as well. Incorrect fundamental knowledge and the assumptions of the science-based first-principles model will lead to inaccurate hybrid model, so it is important for the scientific model to be very accurate. Lack of engineer/scientists having expertise in both domain knowledge and machine learning. Computation infeasibility in certain modeling approaches like inverse modeling.

Data cleaning, preprocessing, feature engineering maybe difficult in certain cases but may be imperative in science-based model parameter estimation hence in these cases the hybrid models may increase the complexity compared to a stand-alone ML models like Neural Network which may not require feature engineering. Model predictions must not only be accurate but also with lower uncertainty which may be difficult for certain hybrid model methods.

There is a lot of scope of using hybrid SGML methodologies in chemical process modeling, summarizing here some of the opportunities and areas where they can be beneficial. As we have seen Hybrid SGML models are useful for extrapolation and predicting beyond operating range, hence it will be particularly useful for processes development. Process fault diagnosis and anomaly detection is one such area where data-based methods have been used extensively, thus there is opportunity to combine scientific knowledge as well to make the anomaly detection process more scientifically consistent.

Table 1 summarizes all the hybrid SGML models and their requirements, advantages, limitations and potential applications.

Table 1. Summary of hybrid SGML approach

| Hybrid SGML Modeling | | Science-based model /knowledge | ML model | Advantages | Limitations | Potential applications |
|---|---|---|---|---|---|---|
| **ML compliments science (base model: science-based)** | | | | | | |
| Direct Hybrid modeling | Series | Science-based model (SBM) | Regression | Extrapolation; Parameter estimation; data augmentation | Limited by data for parameter estimation, Interpolation, Scientific Knowledge dominated | Kinetic estimation [77-78] Soft sensor [86] Process optimization [75] Process design [51] Process modeling [29,76] |
| | Parallel | SBM | Regression | Improved accuracy of prediction, Interpolation | Scientific consistency depends on SBM, Extrapolation; Data-dominated | Process scale-up[41] Process optimization [33,68] Process control [36,71-72] Soft sensor [33] Process monitoring [68] Predictive maintenance] [66,74] |
| | Series-parallel | SBM | Regression | Higher Accuracy, Interpolation | Increased model complexity, Data-dominated | Process optimization [79,80] Process monitoring and control [58] Plant-model mismatch [17] |
| Inverse modeling | | SBM | Probabilistic, Regression | Computationally cheaper inverse problem | Lower generality of the model | Product design and development [84-85] Polymer grade change, Material design[87] |
| Reduced-order models | | SBM | Regression | Fast online deployment, reduce model complexity | Higher Bias, limited by SBM accuracy | Process optimization at plant scale [45,94], Dynamic modeling [93,95] Soft sensor [90] Feasibility analysis [91-92] |

| Uncertainty quantification | SBM | Probabilistic | Gives real error estimate and solution space | Limited by SBM assumptions, parameters | Process design and development [97] Reaction kinetics [98] Feasibility analysis [102] |
| --- | --- | --- | --- | --- | --- |
| Discovering scientific law | SBM | Regression, probabilistic | System stability interpretability | Limited by data size/availability | Physical chemistry [105] Fluid dynamics [104] Thermodynamics phase equilibrium[106] |
| **Science compliments ML (base model : ML)** | | | | | |
| Science-guided design | Laws or SBM | Deep neural network (DNN); Neural Diff. Eqn. | Scientifically consistent and interpretable | Requires deep scientific knowledge of system | Dynamic system [109] Process control [110,114], Process monitoring [111] |
| Science-guided learning | Laws | DNN | Scientifically consistent and interpretable | Possible lower prediction accuracy | Process design and development [117-118], Process monitoring, Process flow [119] |
| Science-guided refinement | SBM | Probabilistic | Less effort in feature selection | Limited by SBM assumptions, parameters | Process design, discovery of materials [120-123] |

# 7 | CONCLUSION

We present a broad perspective of hybrid modeling with a science-guided machine learning (SGML) approach and its application in bioprocessing and chemical engineering. We give a detailed review and exposition of the hybrid SGML modeling approach and its applications, and classify the approach into two categories. The first refers to the case where a data-based ML model compliments and makes the first-principles science-based model more accurate in prediction, and the second corresponds to the case where scientific knowledge helps make the ML model more scientifically consistent. We point out some of the areas of SGML which have not been explored much in chemical process modeling and have potential for further use like in the areas where Science can help improve the data-based model by improving the model design, learning and refinement. We also illustrate some of these applications of the hybrid SGML methodologies for industrial polymer/chemical process improvement.

Thus, based on our review, we recommend that the use of hybrid models will perform better than standalone ML for applications like process development, since they are better at extrapolation while standalone ML models which can be adequate for prediction in a steady running plant.

**Acknowledgement**

We gratefully acknowledge Aspen Technology, Inc., for their support of the Center of Excellence in Process System Engineering in the Department of Chemical Engineering at Virginia Tech since 2002. We thank Mr. Antonio Pietri, CEO, Mr. Willie K. Chan, Chief Technilogy Officer, Dr. Steven Qi, Vice President for Training and Customer Support, and Mr. Daniel Clenzi, Director of University Programs of Aspen Technology, Inc. for their strong support. We would like to thank Professor Anuj Karpatne for sharing his knowledge of theory-guided data science through his course, physics-guided machine learning, at Virignia Tech.

17. Chen Y, Ierapetritou M. A framework of hybrid model development with identification of plant-model mismatch. AIChE Journal. 2020;66(10):e16996.

18. Von Stosch M, Oliveira R, Peres J, de Azevedo SF. Hybrid semi-parametric modeling in process systems engineering: Past, present and future. Computers & Chemical Engineering. 2014;60:86-101.

19. Qin SJ, Chiang LH. Advances and opportunities in machine learning for process data analytics. Computers & Chemical Engineering. 2019;126:465-73.

20. Qin SJ, Guo S, Li Z, Chiang LH, Castillo I, Braun B, Wang Z. Integration of process knowledge and statistical learning for the Dow data challenge problem. Computers & Chemical Engineering. 2021; 153:107451.

21. O'Brien CM, Zhang Q, Daoutidis P, Hu WS. A hybrid mechanistic-empirical model for in silico mammalian cell bioprocess simulation. Metabolic Engineering. 2021; 66:31-40.

22. Pinto J, de Azevedo CR, Oliveira R, von Stosch M. A bootstrap-aggregated hybrid semi-parametric modeling framework for bioprocess development. Bioprocess and Biosystems Engineering. 2019 ;42(11):1853-65.

23. Chopda V, Gyorgypal A, Yang O, Singh R, Ramachandran R, Zhang H, Tsilomelekis G, Chundawat SP, Ierapetritou MG. Recent advances in integrated process analytical techniques, modeling, and control strategies to enable continuous biomanufacturing of monoclonal antibodies. Journal of Chemical Technology & Biotechnology. 2021; http:/doi.org/10.1002/jctb.6765

32. Bellos GD, Kallinikos LE, Gounaris CE, Papayannakos NG. Modelling of the performance of industrial HDS reactors using a hybrid neural network approach. Chemical Engineering and Processing: Process Intensification. 2005;44(5):505-15.

33. Chang JS, Lu SC, Chiu YL. Dynamic modeling of batch polymerization reactors via the hybrid neural-network rate-function approach. Chemical Engineering Journal. 2007;130(1):19-28.

34. Hinchliffe M, Montague G, Willis M, Burke A. Hybrid approach to modeling an industrial polyethylene process. AIChE Journal. 2003;49(12):3127-37.

35. Madar J, Abonyi J, Szeifert F. Feedback linearizing control using hybrid neural networks identified by sensitivity approach. Engineering Applications of Artificial Intelligence. 2005;18(3):343-51.

36. Simutis R, Lübbert A. Hybrid approach to state estimation for bioprocess control. Bioengineering. 2017;4(1):21-9.

37. Cubillos F, Callejas H, Lima EL, Vega MP. Adaptive control using a hybrid-neural model: application to a polymerisation reactor. Brazilian Journal of Chemical Engineering. 2001;18(1):113-20.

38. Doyle III FJ, Harrison CA, Crowley TJ. Hybrid model-based approach to batch-to-batch control of particle size distribution in emulsion polymerization. Computers & Chemical Engineering. 2003;27(8-9):1153-63.

39. Rodriguez-Granrose D, Jones A, Loftus H, Tandeski T, Heaton W, Foley KT, Silverman L. Design of experiment (DOE) applied to artificial neural network architecture enables